\documentclass[sigconf]{acmart}
\AtBeginDocument{%
  \providecommand\BibTeX{{%
    \normalfont B\kern-0.5em{\scshape i\kern-0.25em b}\kern-0.8em\TeX}}}

\renewcommand\footnotetextcopyrightpermission[1]{} 
\begin{document}
\title{\textbf{$\pi_t$} - Enhancing the Precision of Eye Tracking using Iris Feature Motion Vectors
}

\author{Aayush K. Chaudhary}
\email{akc5959@rit.edu}
\affiliation{%
  \institution{Rochester Institute of Technology}
  \city{Rochester}
  \state{NY}
  \country{USA}}

\author{Jeff B. Pelz}
\email{pelz@cis.rit.edu}
\affiliation{%
  \institution{Rochester Institute of Technology}
  \city{Rochester}
  \state{NY}
  \country{USA}}

\begin{abstract}
A new high-precision eye-tracking method has been demonstrated recently by tracking the motion of iris features rather than by exploiting pupil edges. While the method provides high precision, it suffers from temporal drift, an inability to track across blinks, and loss of texture matches in the presence of motion blur. In this work, we present a new methodology ($\pi_t$) to address these issues by optimally combining the information from both iris textures and pupil edges. With this method, we show an improvement in precision (S2S-RMS \& STD) of at least 48\% and 10\% respectively while fixating a series of small targets and following a smoothly moving target. Further, we demonstrate the capability in the identification of microsaccades between targets separated by $0.2^{\circ}$.
\keywords{Eye tracking methodology, iris features, eye movements, video-based eye tracking, iris segmentation, fixations, smooth pursuit, microsaccades, gaze estimation.}
\end{abstract}
\maketitle

\section{Introduction}
Estimation of gaze in current non-invasive, video-based eye trackers allows research involving visual perception, eye movement studies, virtual reality, and augmented reality in natural conditions ~\citep{li2008model,Kassner:2014:POS:2638728.2641695, san2010evaluation,holmdahl2015build,pupil2019invisible}. Current feature-based eye-trackers mainly estimate gaze by fitting a mathematical model (model-based or regression-based in 2D or 3D) to localized eye features such as the pupil, iris border, and corneal reflection (CR)~\citep{tsukada2011illumination,kassner2014pupil,dierkes2018novel,li2005starburst,hansen2009eye}.  

The spatial quality of the estimated gaze is characterized mainly by two widely used metrics; accuracy and precision. The degree to which an eye tracker can estimate the correct gaze position for a known target is referred to as its \textit{accuracy}, whereas the degree of spatial and temporal dispersion of gaze during fixations is referred to as its \textit{precision}~\citep[p.~182]{edlund2019advanced}. Various factors such as the inability of the participant to fixate at the desired target, the video quality of an eye tracker (resolution and compression artifacts), the experimental setup (on-axis or off-axis), calibration decay and the gaze-estimation algorithms limit the precision and accuracy of these systems~\citep{blignaut2019cost,ehinger2019new}.

Among those factors, the selection of a gaze-estimation algorithm can influence the reported gaze significantly~\citep{hansen2010eye, villanueva2008evaluation}. Current video-based eye trackers are limited by signal noise due to reliance on localizing the edges of the pupil/iris boundary. Low-level image features such as edges are intolerant to illumination changes, occlusion caused by eyelashes/eyelids, CR at the pupil boundaries, etc. They are also highly dependent on parameters such as threshold values. Further, when eye-cameras are off-axis, the objective function of fitting an ellipse on the projected 2D image of iris or pupil might produce an error as these features are not perfect ellipses~\citep{swirski2015gaze,villanueva2008evaluation,wang2019center}.Recently, convolutional neural network-based approaches (appearance-based models) have been developed~\citep{park2018deep,park2018learning,kim2019nvgaze}, which focus on taking advantage of extensive training sets, learning-based optimization and generalization, preparation of synthetic data with natural features using generative adversarial networks~\citep{kim2019nvgaze,wood2016learning}, etc. However, the gaze estimation results are biased towards the training set and have only reached an accuracy of 2.06 ($\pm0.44$) degrees~\citep{kim2019nvgaze} for real subjects. There is still room for improvement in these approaches before its extensive use in research, especially in tasks requiring high precision.

In another approach,~\cite{pelzwitzner} tracked a large number of \textit{motion vectors} made up of iris features across adjacent frames in order to extract the velocity of the eye over time, integrating the velocity to obtain eye position. They emphasized improving the temporal precision and accuracy of the eye-tracking system by considering a large number of iris features where the noise was addressed in the spatial domain rather than the temporal domain.~\cite{Chaudhary_Pelz_2019} extended~\cite{pelzwitzner}'s approach to demonstrate a method to achieve a high precision system capable of detecting small eye movements as small as 0.2 degrees with high confidence. While~\cite{pelzwitzner} and~\cite{Chaudhary_Pelz_2019} both show significant gains in precision due to the emphasis on a large number of motion vectors, neither addresses the issue of the drift inherent in a system that determines position purely by the integration of velocity over time. Small errors in approximations of the velocity induced by the use of central tendency metrics like geometric median~\citep{pelzwitzner,Chaudhary_Pelz_2019} accumulate over time, resulting in temporal drift which degrades accuracy in gaze estimations~\citep{chaudhary2019motion}. 

Additionally,~\cite{pelzwitzner} and~\cite{Chaudhary_Pelz_2019} only used trials without blinks because the method relies solely on the integration of iris motion vectors, which are absent during blinks. If gaze position changes while the eyelid covers the iris features, gaze position based on integrated motion values are inaccurate. Motion blur can also degrade performance as relatively few matches are available.

This raises a concern regarding the overall concept of relying completely on the integration of velocity for the position signal, especially during rapid motion and pupil occlusion. One alternative approach might be to use the velocity integration method to provide only relative position information and use it in parallel with a traditional eye-position signal (e.g., P-CR or 3D gaze vector). In essence, such a system would provide high precision data during fixations and smooth pursuit, but rely on traditional methods at all other times.

The goal of this paper is to address these issues, namely erroneous drifts, handling of blinks, and motion blur. Our approach is to use traditional approaches such as (P-CR) based system with an iris-based velocity system as proposed in~\citep{pelzwitzner,Chaudhary_Pelz_2019}.  P-CR based gaze estimation has been used extensively since the 1960s and is still used today in some commercial eye-trackers. However, precision is constrained because of reliance on pupil edges and the bright corneal reflection~\citep{li2008model}. By optimally combining the \textbf{\textit{P}}-CR \textit{position} with the \textbf{\textit{i}}ris \textit{velocity} (computed from \textbf{\textit{t}}extures) [``$Pi_t$''  (or ``$\pi_t$'')] we demonstrate a hybrid method with high precision without the issues caused by drift, blinks, or motion blur. 

Observations made from multiple sources (e.g., pupil and iris estimates from different measurement techniques) can be fused with various state estimation techniques. Maximum likelihood and maximum posterior, the Kalman filter~\citep{kalman1960new}, and the particle filter~\citep{del1996non} are some of the common state estimation techniques used based on linear/non-linear dynamic measurements systems~\citep{castanedo2013review}. The disadvantages of maximum likelihood and maximum posterior are that in order to reduce the bias of the solution, they require an empirical model of the sensor with a possibly large number of samples~\citep{castanedo2013review}. As our data can be assumed to be from linear sources with Gaussian distributed noise (random variable derived by a large number of identically independent iris feature matches/pupil edges), modeling with the Kalman filter is a good fit as it assures the optimal estimation on this type of data~\citep{alofi2017review}. The use of the particle filters is sophisticated and well-practiced for non-linear data, but to obtain small variance in the estimations requires a relatively large data sample~\citep{alofi2017review}.

The Kalman filter is designed to combine uncertain information from multiple independent sources to estimate a more confident (certain) approximation~\citep{pei2017elementary}. Our goal is to combine the information derived from the iris (\textit{I}: high precision but possibly contaminated with drift, motion blur, or blinks) with the information from the pupil (\textit{P}: no drift but noisy) by considering a convex combination of the two estimates with a general form  $\beta \times I + (1-\beta) \times P $, where  $(0 \leq \beta \leq 1)$~\citep{pei2017elementary}. $I$ and $P$ are derived in Section ~\ref{sec:model_formulate} and represent iris and pupil information, respectively, and $\beta$ is scaled with confidence in the information from the iris. Note that position information (P) is computed independently for each frame. As a result there is no accumulated drift, but the signal is noisy because of reliance on edge/boundary calculations based on a small number of features (edges) on each video frame. A position signal can also be computed by integrating the velocity signal derived from the iris features in consecutive frames. That position signal, however, suffers from drift over time because even small errors are compound over large numbers of frames.

Here, combining the information from the two signals generated from two independent methods is not straightforward. We are only interested in the position information from the pupil center, and only the velocity information from the iris features so that the noise (from the pupil signal) and the drift (from iris signal) are not considered. Our work incorporates the two signals in the different domains that are related by first-order derivative/integration, which can further be applied to any application which has a spatial drift over time. We demonstrate this strategy in the field of eye-tracking to devise a post-hoc evaluation pipeline on a chin-rest system with both high accuracy and high precision for stimuli displayed in an automated teleprompter setup.  

The major contributions of this paper are as follows:
\begin{enumerate}
\item We demonstrate a new methodology to improve precision of any pupil detection technique by incorporating information from multiple features of the iris. This is one of the earliest attempts to combine regression-based approaches (P-CR) with computer-vision approaches such as iris feature matching and tracking. This paper opens an interesting area to be explored i.e. combining information from any eye-trackers (P-CR, 3D based approaches, appearance-based models) with iris-based estimates.
\item We present a modified Kalman filter approach that helps to estimate a reliable signal by disentangling useful pieces of information from two independent sources (one precise, but drifting signal and another accurate but noisy signal).
\item With the off-the-shelf components like a digital camera, synchronized displays and IRLEDs, we show sample-to-sample root mean square error (S2S-RMS) of ~$0.05^{\circ}$ for real human-eye.
\end{enumerate}

\section{Literature Review}
Depending on the use of either polynomial gaze function or fitting a 3D model of the eye to predict gaze, video-based eye-tracking methodology is broadly divided into two categories; regression-based and model-based gaze estimation~\citep{hansen2009eye}. Both categories are reliant on eye features such as pupil center, pupil/iris contours, glints (CR), facial features~\citep{chen2008robust}, limbus, eye-corners, iris features, etc.~\citep{tsukada2011illumination,li2005starburst,park2018learning,pelzwitzner,Chaudhary_Pelz_2019,hansen2009eye,chen2008robust,swirski2012robust}. Regression-based gaze estimation uses a mapping function to predict 2D gaze coordinates based on the features. In contrast, model-based approaches allow gaze estimation in 3D space. Model-based approaches such as~\citep{tsukada2011illumination,swirski2012robust,shih2004novel,guestrin2006general,chen20083d} assume a spherical eye model and approximate a gaze direction based on  pupil/iris contours. These models approximate the solution based on several assumptions such as neglecting perspective projection and the effects of refraction in~\citep{shih2004novel}, virtual pupil lying in the optical axis in~\citep{chen20083d} and neglecting corneal reflection in~\citep{swirski2012robust}. Moreover, recent publications~\citep{dierkes2018novel,dierkes2019fast}  have also accounted for the corneal reflection in the 3D-model. 


Recently, appearance-based models have appeared in the field of eye-tracking, which try to learn a direct mapping between the images and the gaze direction~\citep{park2018deep,park2018learning}. These models have been shown to perform better than the above models in person-independent gaze estimation and unconstrained environments but are particularly biased towards the training set. Other challenges for appearance-based models are finding ways to incorporate prior knowledge in a differential manner, the need for enormous labeled data sets~\citep{marcus2018deep}, computational cost, and the difficulty of understanding what the models have actually 'learned'~\citep{park2018learning}. Even though appearance-based models were incorporated with additional regression-based or model-based methods in~\citep{park2018deep,park2018learning}, the method is still not reliable for high precision tasks.

We propose a regression-based gaze estimation method that has the potential to be further modified for model-based gaze estimation. Our priority is to achieve a high-precision system and are focused on tracking features that are more reliable for that purpose. To do so, we are extending~\citep{Chaudhary_Pelz_2019} previous work, which demonstrated a high-precision task (microsaccade detection) by integrating eye velocity computed from a large number of iris features. To address the issues existing in that method, we enhance that method by optimally adding pupil-center tracking. Pupil location has been tracked in numerous ways including the Starburst algorithm (based on the iterative selection of candidate points obtained by ray following to find best-fit)~\citep{li2005starburst}, an image-aware support function to fit an ellipse to the pupil edge~\citep{swirski2012robust}, and fitting an ellipse to connected components~\citep{Kassner:2014:POS:2638728.2641695}.  In this work, we deploy the method of~\cite{Kassner:2014:POS:2638728.2641695} to detect the pupil center, but we capture a high-resolution image of the eye as in~\citep{Chaudhary_Pelz_2019}. Our regression-based gaze estimation method is based on those signals.

\section{Methods}
\begin{figure}
\begin{center}
\includegraphics[width=\linewidth]{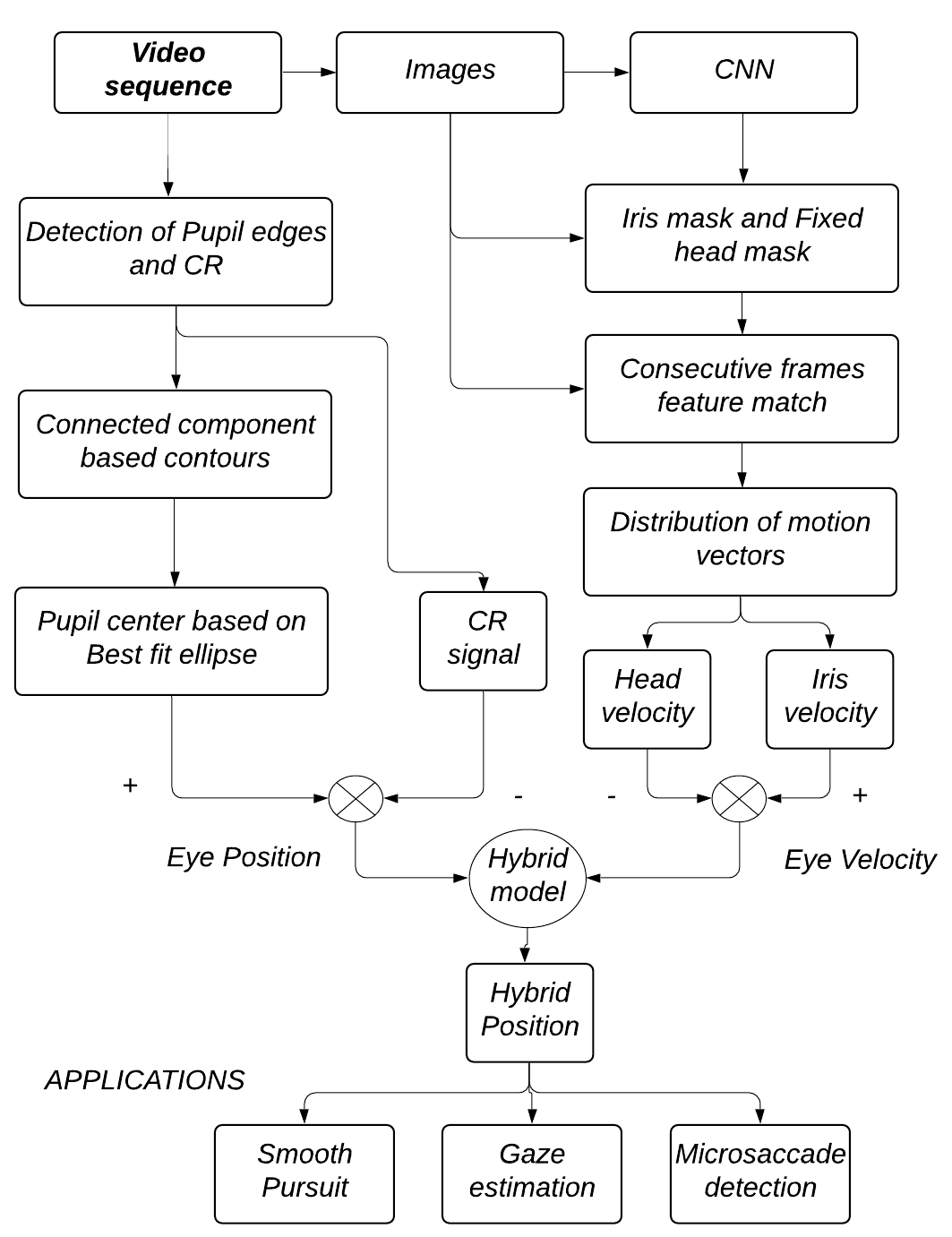}
\end{center}
\caption{Basic flow of our methodology.}
\label{fig:block}
\end{figure}

This section formulates the problem of estimating an approximate hybrid position ($\pi_t$) by extracting the P-CR relative eye position and iris velocity (\textit{i}). The detailed block diagram is shown in Figure~\ref{fig:block}. Frames are extracted from the video sequence and fed to specific blocks for processing. In one of the blocks, images are fed to a CNN to obtain an iris mask, which is a crucial step to define the region of interest for another step whose objective is to approximate the iris velocity based on the motion distribution of feature matches in consecutive frames. The next step is the determination of the pupil center based on ellipse fitting, as proposed in~\cite{Kassner:2014:POS:2638728.2641695}. To compensate for head movements, we determine the average CR signal of multiple glints as well as approximate head movement velocity based on a fixed head mask on the subject. Based on these signals, $\pi_t$ is computed and demonstrated in applications such as gaze estimation, smooth pursuit analysis, and microsaccade detection.

\begin{figure*}
\begin{center}
\includegraphics[width=\linewidth]{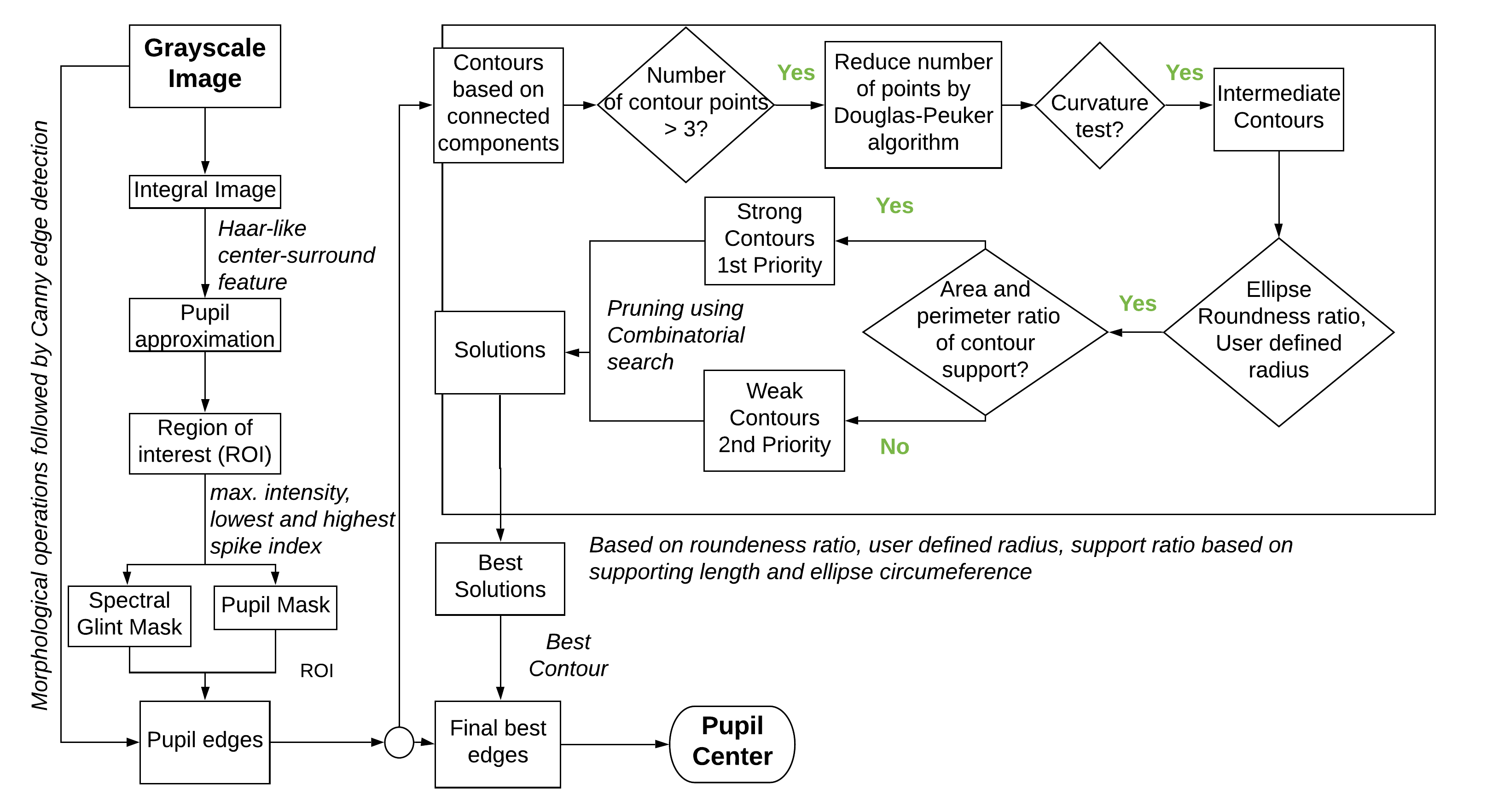}
\end{center}
\caption{2D pupil center detection based on open-source of Pupil Capture~\citep{Kassner:2014:POS:2638728.2641695}}
\label{fig:pupillabs}
\end{figure*}

\subsection{Pupil position}
The pupil center (P) is specified as the ellipse center given by the 2D pupil detection model of the Pupil Labs software of~\cite{Kassner:2014:POS:2638728.2641695}.~\cite{Kassner:2014:POS:2638728.2641695} initially compute an integral image from the grayscale image and find an approximation to the pupil and region of interest (ROI) for pupil detection based on response to Haar-like center-surround features~\citep{swirski2012robust}. 

In the obtained ROI, spectral glint masks and a dark pupil mask are created based on the maximum intensity and the lowest and highest spike index. Morphological operations (opening with an elliptical structured kernel of size 9) followed by the Canny Edge detector~\citep{canny1986computational} are then used to find the pupil edges.

In~\cite{Kassner:2014:POS:2638728.2641695}, the contours are found based on connected components of the edges. On those contours which have a minimum size of three points, the number of points that compose that line segment is reduced to find a similar curve using the OpenCV~\citep{bradski2000opencv} implementation of the Douglas-Peuker algorithm~\citep{douglas1973algorithms}. The extracted contour points are then passed through a curvature test such that any three adjacent contour points must have a curvature more than 80 degrees for the curvature to be maintained.

Thus, on the intermediate contours based on continuous curvature, \textit{weak} and \textit{strong} contours are extracted based on parameters such as roundness ratio (ratio of minor and major axis) and user-defined radius limits of ellipse fit. \textit{Strong} contours must also satisfy ellipse criteria for area and perimeter ratio of contour support, whereas weak contours need not. For those contours, either \textit{strong} or \textit{weak}, pruning using an augmented combinatorial search is done in order to find the \textit{solutions}.

\textit{Best solutions} out of these \textit{solutions} are found based on the roundness ratio, the user-defined radius of ellipse fit (70, 200), and support ratio based on supporting edge length and the ellipse circumference. The \textit{best contour} is obtained from the \textit{best solution}, which gives us the final \textit{best edges} (intersection of best contour with initial edges). The final \textit{best ellipse} is based on the ellipse fit on those \textit{final edges} when it satisfies all the above criteria. The center of this final \textit{best ellipse} fit is considered as the pupil-center (P) in this paper. These steps are shown in Figure~\ref{fig:pupillabs}.

\subsection{Iris velocity}
Eye velocity is determined by tracking multiple feature points on the iris, so the first step is defining the iris ROI on the eye images. Semantic segmentation models like U-Net~\citep{ronneberger2015u} and RITnet~\citep{9022181} can be used to extract the segmented iris mask from a given eye image.  We adopt the U-Net architecture as in~\citep{Chaudhary_Pelz_2019} with a modification during model training. Instead of reshaping the 960$\times$540 to 224$\times$224 as in~\citep{Chaudhary_Pelz_2019}, we initially partition each eye image to 540$\times$540 before resizing to 224$\times$224. This maintains the aspect ratio of the image and eliminates unnecessary pixels as the eye size $<540$ pixels. Then, we follow the same steps as in~\citep{Chaudhary_Pelz_2019} on the segmented mask, i.e., we extract iris features from a Contrast Limited Adaptive Histogram equalized~\citep{pizer1987adaptive} grayscale image. These features are matched in consecutive frames based on the Lowe's ratio distance test followed by RANSAC. These matched features represent the iris feature movement vector. Final iris velocity is computed by calculating the geometric median of these movement vectors~\citep{Chaudhary_Pelz_2019}.

\subsection{Problem Formulation}
\label{sec:model_formulate}
We have two sources of information: iris \textit{velocity} and pupil \textit{position} which we intend to combine with the Kalman filter, an optimal estimation technique for linear systems with Gaussian error~\citep{anderson2012optimal}. A straightforward interpretation of the Kalman filter update equations is that they scale measurements from two sources by their corresponding precision matrix (the inverse of the covariance matrix) and then take the weighted sum as shown in Equation ~\ref{eq:basic}.
\begin{equation}
\label{eq:basic}
\overline{H}=\Sigma(\beta*P+(1-\beta)*I) 
\end{equation}
where $P$, $I$ and $H$ are the pupil position, the iris position and the hybrid position respectively.

In our case, one source of measurement is the pupil position, and the other is the iris velocity (which is integrated to compute the iris position). The information in iris velocity can only provide us information about the iris position up to a constant bias value. If we use a traditional update of the Kalman filter (Equation \eqref{eq:basic}), we would be using incorrect bias induced by the integration of iris velocity, and error would still creep into our estimated solution. 
The estimated signal would also drift over time, primarily affected by the bias information from the iris velocity measurement. Therefore, while combining iris velocity measurement with the pupil position measurement, we must ignore any bias information obtained from the iris velocity measurement to gain the benefit in precision from the velocity measure without degrading accuracy.

Therefore, we need to make some crucial changes to the Kalman filter approach. To do so, we take the probabilistic interpretation of the Kalman filter. Using a probabilistic framework helps us combine two information sources from different domains in the same spirit of the Kalman filter. In a probabilistic interpretation, a Kalman update equation is interpreted as the posterior mean in which prior and likelihood are both linear. In this framework, we can easily integrate two measurements from different domains by defining a prior distribution which behaves like a Gaussian in a gradient domain, i.e., whose derivative is Gaussian. Using this approach, we can derive a Kalman update equation (as in~\cite{ghimire2019noninvasive}).

We can fuse the information of the pupil position ($P$) with the iris velocity ($i$ or $\textbf{D}I$) to obtain the hybrid position (H), where \textbf{D} is the spatial gradient operator. In general, if $\textbf{P}(P|H)$ is the likelihood, $\textbf{P}(H|I)$ refers to the prior probability distribution, and $\textbf{P}(H|P,I)$ refers to the posterior distribution, then the posterior probability can be computed as
\begin{equation*}
\begin{split}
 \textbf{P}(H|P,I) &=\frac{\textbf{P}(P|H)\textbf{P}(I|H) \textbf{P}(H)}{\textbf{P}(P,I)} 
=\frac{ \textbf{P}(P|H) \textbf{P}(H|I)}{\textbf{P}(P|I)}  \\
\end{split}
\end{equation*}
where $\textbf{P}(P|I)$ is the normalization factor. So, we have $\textbf{P}(H|P,I)\propto{\textbf{P}(P|H) \textbf{P}(H|I)}$. Here, $\textbf{P}(P|H)$ and $\textbf{P}(H|I)$ are defined as:

\begin{equation*}
\textbf{P}(P|H)=\frac{\exp^{-\frac{(P-H)^2}{2\sigma_P^2}}}{\sqrt{2\pi\sigma_P^2}}=\frac{\exp^{-(P-H)^T\beta_P\textbf{I}(P-H)/2}}{\sqrt{2\pi\sigma_P^2}}
\end{equation*}

\begin{equation*}
\begin{split}
\textbf{P}(H|I)&=\frac{\exp^{-\frac{(\frac{\textbf{D}H}{\textbf{D}t}-\frac{\textbf{D}I}{\textbf{D}t})^2}{2\sigma_I^2}}}{\sqrt{2\pi\sigma_I^2}} 
=\frac{\exp^{-{({H}-I)^T\beta_I\textbf{D}^T\textbf{D}({H}-I)}/2}}{\sqrt{2\pi\sigma_I^2}}
\end{split}
\end{equation*}
where $\beta_I=1/{\sigma_I^2}$ and $\beta_P=1/{\sigma_P^2}$. Higher values of $\beta$ indicate lower standard deviation and more certainty.

Note that we have used the spatial gradient of the iris position. For a prior distribution, the low dimensional structure of the signal can be considered to model the prediction error~\citep{ghimire2019noninvasive}. In our case, we find the hybrid signal by minimizing the mean square estimation between hybrid velocity and iris velocity and also between hybrid position and pupil position. 

An important property of Gaussian distributions is that the product of two Gaussian distributions is a Gaussian distribution~\citep[p.~638]{bishop2006pattern}. Thus,\\
\begin{equation*}
\textbf{P}(H|P,I)={\textbf{P}(P|H) \textbf{P}(H|I)}
\end{equation*}

\begin{equation}\label{eq:appendix_elaborate}
=\frac{\exp^{-(P-H)^T\beta_P\textbf{I}(P-H)/2}}{\sqrt{2\pi\sigma_P^2}}\frac{\exp^{-{({H}-I)^T\beta_I\textbf{D}^T\textbf{D}({H}-I)}/2}}{\sqrt{2\pi\sigma_I^2}}
\end{equation}
The equation is simplified in Appendix A. Further our term $\textbf{P}(H|P,I)$, can be expressed as 

\begin{equation}\label{eq:meanestimate}
\begin{split}
\textbf{P}(H|P,I)&=K{\exp^{-1/2{(H-\overline{H})^T\Sigma^{-1}(H-\overline{H})}}} \\
&=K{\exp^{-1/2{(H^T\Sigma^{-1}H-H^T\Sigma^{-1}\overline{H}+...)}}}
\end{split}
\end{equation}

Comparing, Appendix A (Equation~\ref{eq:longe}) and ~\ref{eq:meanestimate}, we get the mean estimate ($\overline{H}$) and the covariance ($\Sigma$) of the hybrid position~\citep[p.~639]{bishop2006pattern} as 
\begin{equation*}
{\Sigma}^{-1}=(\beta_P\textbf{I}+\beta_I\textbf{D}^T\textbf{D})\label{eq:inv_cov}
\end{equation*}
\begin{equation}
{\Sigma}=(\beta_P\textbf{I}+\beta_I\textbf{D}^T\textbf{D})^{-1}
\label{eq:3conv}
\end{equation}
\begin{equation*}
\Sigma^{-1}\overline{H}=\beta_P\textbf{I}P+\beta_I\textbf{D}^T\textbf{D}I
\end{equation*}
\begin{equation}
\overline{H}=\Sigma(\beta_P\textbf{I}P+\beta_I\textbf{D}^T\textbf{D}I)\label{eq:main2}
\end{equation}

It is important to note that $\textbf{D}^T\textbf{D}$ is a non-invertible matrix. Therefore as the value of $\beta_P$ tends towards 0, the determinant of  $\Sigma^{-1}$ approaches 0, as shown in Appendix B. The inverse operation cannot be obtained for the covariance matrix. So it is preferable to keep the value of $\beta_P$ non-zero as it is the limiting case for our approach.

With this update strategy, the bias information in the velocity measurement does not influence the solution. Hence, the solution correctly uses the rest of the information in the velocity to give a reasonable estimate of the position. This setup allows handling the drift in the estimated signal, and essential information from the iris velocity is preserved.

Additionally, the modifications made to the Kalman filter also handles latency issues (discussed further in Section~\ref{qualitative}) in the traditional Kalman filter-based approach. Rather than a traditional filtering approach, the formulation we use helps in estimation by combining different temporal frequency components from the pupil and iris signals. Low-temporal frequency components of the pupil are combined with the high-temporal frequency components from the iris. This combination of frequency components results in no time-lag (run-time). Note that the addition of high-frequency components is essential as this data mainly carries information about tremors and microsaccades, which are crucial for high precision tasks.

\subsection{Positional Difference Compensation}
The pupil position and the iris velocity in the above relations are extracted without taking into account the movements of the eye camera (headgear slippage and relative camera movement)~\citep{li2008model}. To compensate for error induced by these events, we compute the P-CR vector to find the relative position of the eye to the camera. Additionally, we compute the head movement velocity to the camera (head velocity: \textbf{D}$H_v$) for a user-defined ROI on the forehead using a similar approach as used for computing iris velocity. Hence, in our case for the hybrid model, the gaze vector is computed when the relative position between the principal eye signal (iris or pupil) is subtracted with a compensatory movement (CR or $H_v$). So, the overall model can be computed as
\begin{equation}
\pi_t=Pi_t=\overline{h}=\Sigma(\beta_P\textbf{I}(P-CR)+\beta_I\textbf{D}^T\textbf{D}(I-H_v))\label{eq:1}
\end{equation}

\begin{figure}
\begin{center}
\includegraphics[width=\linewidth]{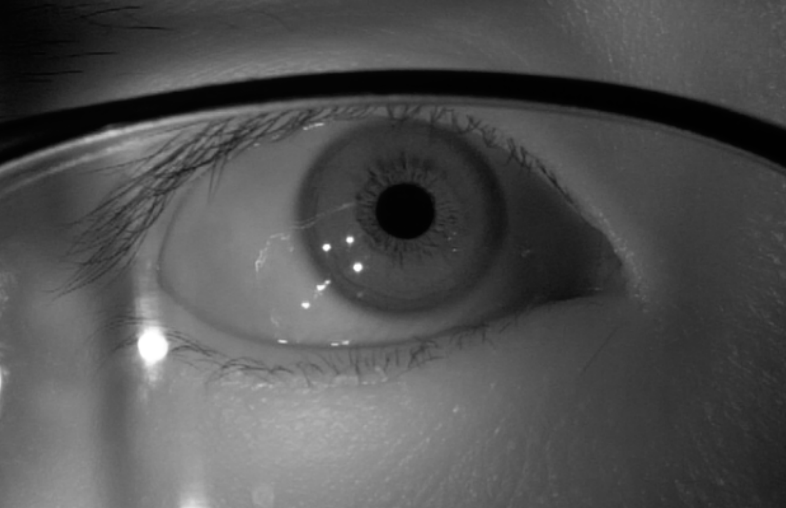}
\end{center}
\caption{Example where one of the four CRs rolls off the cornea onto the sclera.}
\label{fig:roll_off}
\end{figure}

Note that for the CR, we initially segment the iris mask as in ~\citep{Chaudhary_Pelz_2019} to identify the region where the CR is most likely to be present. In the iris segmented region, we identify the bright spots in the mask with an empirically derived hard threshold pixel value of above 140 (for an 8-bit image). For our experiment with four IRLEDs, we initially find the largest bright spot. A small window (ellipse or rectangle depending on the number of points in the contour) around the largest spot is then used to find the remaining valid CRs. The LEDs are placed such that CR usually falls on the iris. However, on infrequent occasions like that seen in Figure ~\ref{fig:roll_off}, one of the CR falls on the sclera. We ignore this CR as a detection but consider all other CRs as valid. So, on those valid CRs, we find the center of the glint using moments. The centers of the detected CRs are used to approximate a circle. The center of this circle is referred to as CR in the following sections.
   
\subsection{Blink Classification and Poor Feature Match}
\label{sec:blinks}
The hybrid model ($\pi_t$) has weighting parameters $\beta_p$ and $\beta_i$, related to confidence in the pupil position and iris velocity, respectively. Except during a blink, they sum to 1. 
The weight $\beta_i$ is a function of the number of detected iris feature matches;  $\beta_i = min(0.9,0.9*(n_{\mathrm{matches}})/(min_{\mathrm{matches}}))$ , where $min_{\mathrm{matches}}$ is a user-set parameter set. The confidence in iris velocity will be degraded if fewer than $min_{\mathrm{matches}}$ feature points are computed, such as in cases of a large saccadic movement, motion blur, and significant compression artifacts. To avoid abrupt changes in $\beta_i$ (from 0.9 to $<$0.9) between two consecutive timestamps when there are only a few feature matches, we use a linear decay function that takes into account the previous/next two timestamp's $\beta_i$ values. 

To classify blinks in the pupil signal, we use a confidence value provided by the open-source Pupil Capture software~\citep{Kassner:2014:POS:2638728.2641695}. While pupil detection confidence decreases even in cases of a `partial blink' where the portion of the pupil is occluded, a few iris feature matches still exist during the partial blink. Thus we classify blinks for pupil and iris separately and set values of $\beta_p$ and $\beta_i$ to 0 only in the case of a complete blink. In the case where pupil confidence is less than user defined confidence threshold, but we still have a few iris-feature matches, $\beta_i$ will be low, and the overall confidence in the position will be low.

For this paper, we set the  user-defined parameters $min_{\mathrm{matches}}$ and confidence threshold to be 50 and 0.3 respectively.

\subsection{Gaze Estimation}
Equation \eqref{eq:1} describes the combination of the pupil position and the iris velocity. These two components represent the calibrated gaze position and the calibrated gaze velocity for the pupil and iris, respectively, which we combine to get a hybrid gaze position. This relation is valid for both cyclopean eye (midpoint of two eyes)~\cite{ono1982cyclopean,hering1977theory} and independent individual eyes. Independent analysis of the individual gaze of each eye supports the study of vergence eye movements, which is not possible with cyclopean gaze estimation. Many current video-based eye trackers report cyclopean gaze estimates because the precision and accuracy of current eye trackers is insufficient to estimate depth based on gaze position. The improved hybrid signal for each eye may provide adequate signal quality to estimate a useful vergence signal. 

For computing the calibrated gaze position from the pupil, calibration is performed by a second-order polynomial of the instructed gaze position and its corresponding relative pupil position ~\cite{cerrolaza2012study,swirski2015gaze}. The same calibration routine creates problems for iris gaze position because the iris drifts over time, and additionally, due to possible pupil dilation/constriction and a significant position change of gaze during blinks. So, instead, we use a calibration scheme based on the iris velocity signal; we know the relative distance between calibration target positions, and we can extract the relative distance of iris position across the saccades during calibration, so we can compute a mapping function between them. We integrate velocity from 30 ms before to 30 ms after each saccade that brings gaze to a fixation target. With the calibrated position and velocity gaze components, we compute the hybrid gaze position in terms of visual angle.


\section{Experimental Design}
\begin{table*}[ht]
\caption{Refinements in the hardware-setup compared to~\cite{Chaudhary_Pelz_2019}}
\label{tab:2}       
\begin{tabular}{|c|c|c|}
\hline
 & \textbf{\cite{Chaudhary_Pelz_2019}} & \textbf{Ours}\\
\hline
Mirrorless Camera & Panasonic Lumix DMC-GH4  & Panasonic Lumix DC-GH5S \\
\hline
Light source & Tungsten halogen source with bifurcated fiber optics & 
Lite-On HSDL-4261 870nm IRED.\\
\hline
Chin rest& Yes  &  Yes  \\
\hline
Forehead rest& Yes &  No (extra degree of freedom) \\
\hline
Stimulus & Printed Snellen Chart & Appearing Intermittently (Teleprompter)\\ 
\hline
Frame rate & 96 Hz & 120 Hz\\
\hline 
Synchronization & No (finite interval used to localize eye movements) & Yes (Photo-diode setup)\\
\hline
\end{tabular}
\end{table*}

\subsection {Camera Setup}
We tested our method with images captured with a Panasonic Lumix DC-GH5S mirrorless digital camera (modified by removing the IR-rejection filter) with a Lumix G Vario 14-140mm II ASPH (H-FSA14140) lens set to a focal length of approximately 100 mm and an aperture of F/8. 
Sensitivity to visible wavelengths was blocked with a DHD IR760 high-pass filter ($T_{380-740} < 1\%, \ T_{780-1200} > 85\%$). The camera was set to an ISO value of 800 and the shutter speed to $1/400$ sec. We recorded binocular eye movements at 120 frames per second (fps) at a resolution of 1920 $\times$ 1080, with IPB compression [FHD/8bit/100M/59.94Hz] with a slow-motion effect of 1/2).

\begin{figure}
\begin{center}
\includegraphics[width=\linewidth]{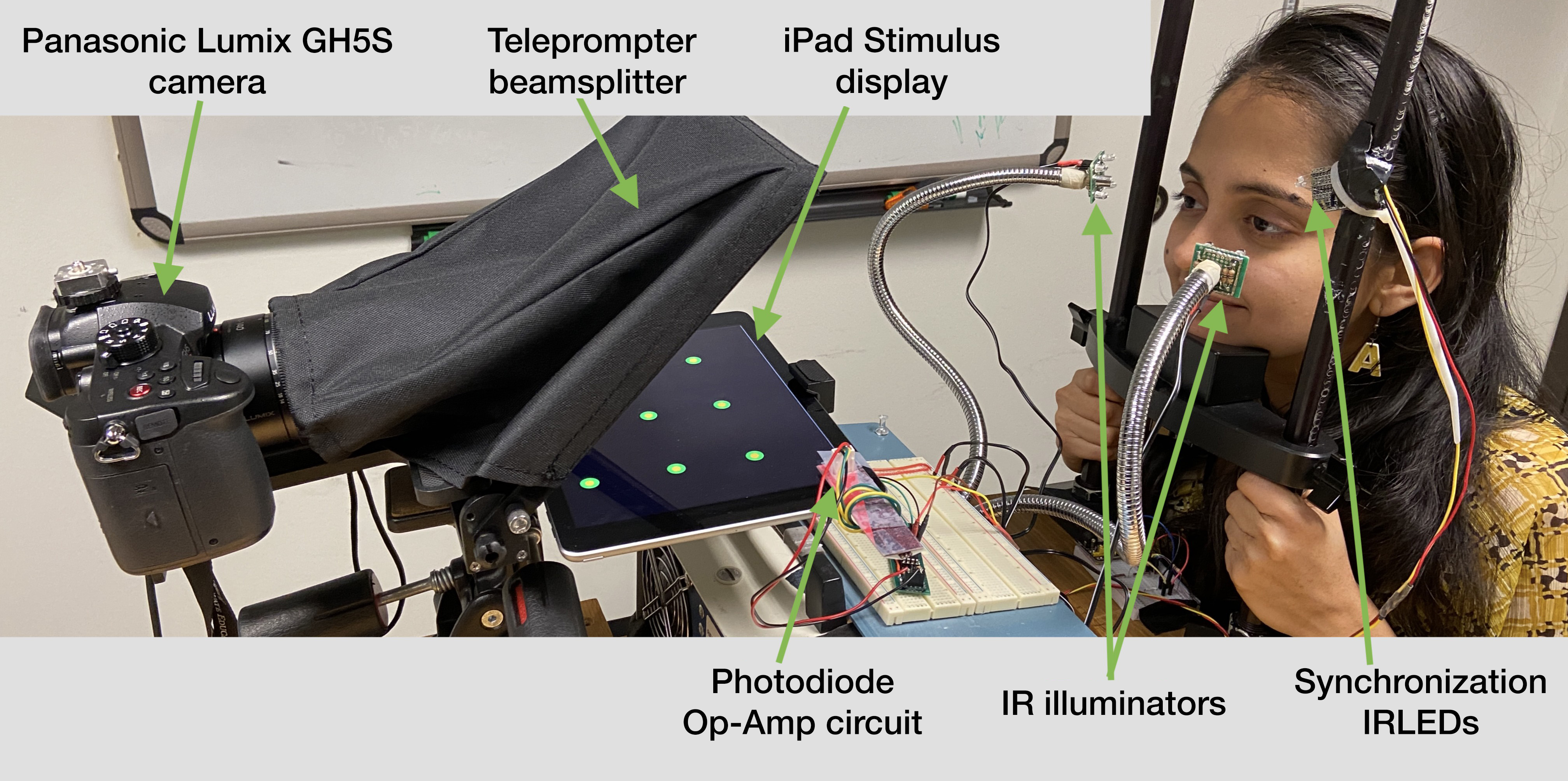}
\end{center}
\caption{Experimental setup with stimulus presented in iPad-teleprompter setup.}
\label{fig:experiment}
\end{figure}

The camera was placed 50 cm from the observers' eyes, whose position was fixed with a UHCOTech HeadSpot chin rest (with no forehead rest). The setup allowed observers to make small rotational and translating head movements without significant distance variation. Four infrared LEDs (IREDs) (HSDL-4261, 870 nm, viewing angle= 26$^{\circ}$), each placed on a square of approximately 2 cm, were used to illuminate each eye uniformly at angle of 25-30 degrees above the horizontal. The IREDs were placed at a distance of approximately 9 cm from the eyes (as shown in Figure~\ref{fig:experiment}) such that an area of approx. 36 $cm^2$ of the eyes was covered to provide proper illumination even if the observer made small head movements. The total irradiance at the eye was 0.01 $W/cm^2$ measured with a calibrated radiometer.

Table ~\ref{tab:2} summarizes the refinements made in the hardware setup compared to~\cite{Chaudhary_Pelz_2019}.

\begin{figure}
\begin{center}
\includegraphics[width=\linewidth]{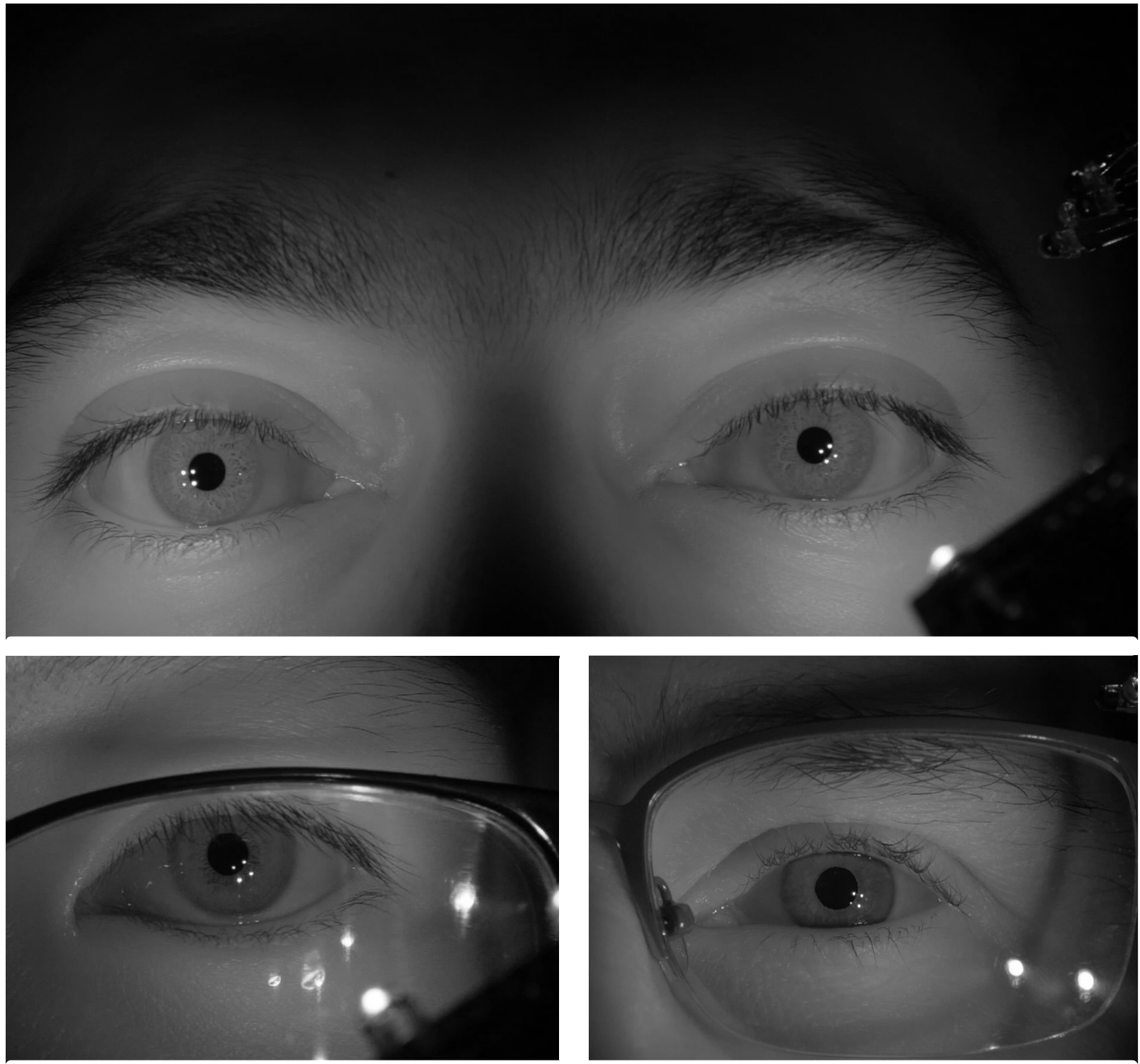}
\end{center}
\caption{Grayscale version of the image captured with the setup (Top row). Bottom row contains cropped version of left eye for different subjects with corrected glasses.}
\label{fig:experiment_images}
\end{figure}

\subsection{Display target}
We used a teleprompter setup (Glide Gear TMP50 20.3 X 17.8 cm) with an iPad (MR7FZLL/A) to display the stimuli. The stimuli were presented as a 30 fps video, the same temporal resolution as the iPad. To synchronize the displayed stimulus, we use a photo-diode setup, as shown in Figure~\ref{fig:experiment}. A region of the video that is not visible in the teleprompter screen contained a unique binary pattern of black and white patches, which transitioned each time the stimulus display changed. The binary pattern was detected by a photo-diode and Op-Amp (LM-358) circuit. Because the iPad LCD display has a faster black-to-white transition than white-to-black, we used the black-to-white transitions to mark events.
To make the display sync signal available in the video record,  the output of the Op-Amp drove a 940 nm IRLED in the field of view of the Lumix camera to indicate when a stimulus was presented to the observer in the video sequence.

Because iPads are raster displays operating at 30 Hz, it takes approximately 33 ms to rewrite the entire display. We derived a parametric model to find the delay based on the display position on the screen. We have accounted for this time delay in our results. The parametric model is given by 
\begin{equation}
   t =(21.4*x)+(4.26*y)-2.35
\end{equation}
where $x$ and $y$ refer to horizontal and vertical stimulus position in pixels, and $t$ is the estimated delay in ms.

\section{Subjects}
We recorded eye movements of seven participants (four males and three females) with a mean age of 31 ($\sigma$=12) with normal or corrected-to-normal (two out of seven subjects) vision. Observers with a varying range of iris pigmentation were selected for the experiment. The experiment was conducted with the approval of the Institutional Review Board, and all participants provided informed consent before starting the experiment.

\section{Tasks}
\begin{figure}[h]
\begin{center}
\includegraphics[width=\linewidth]{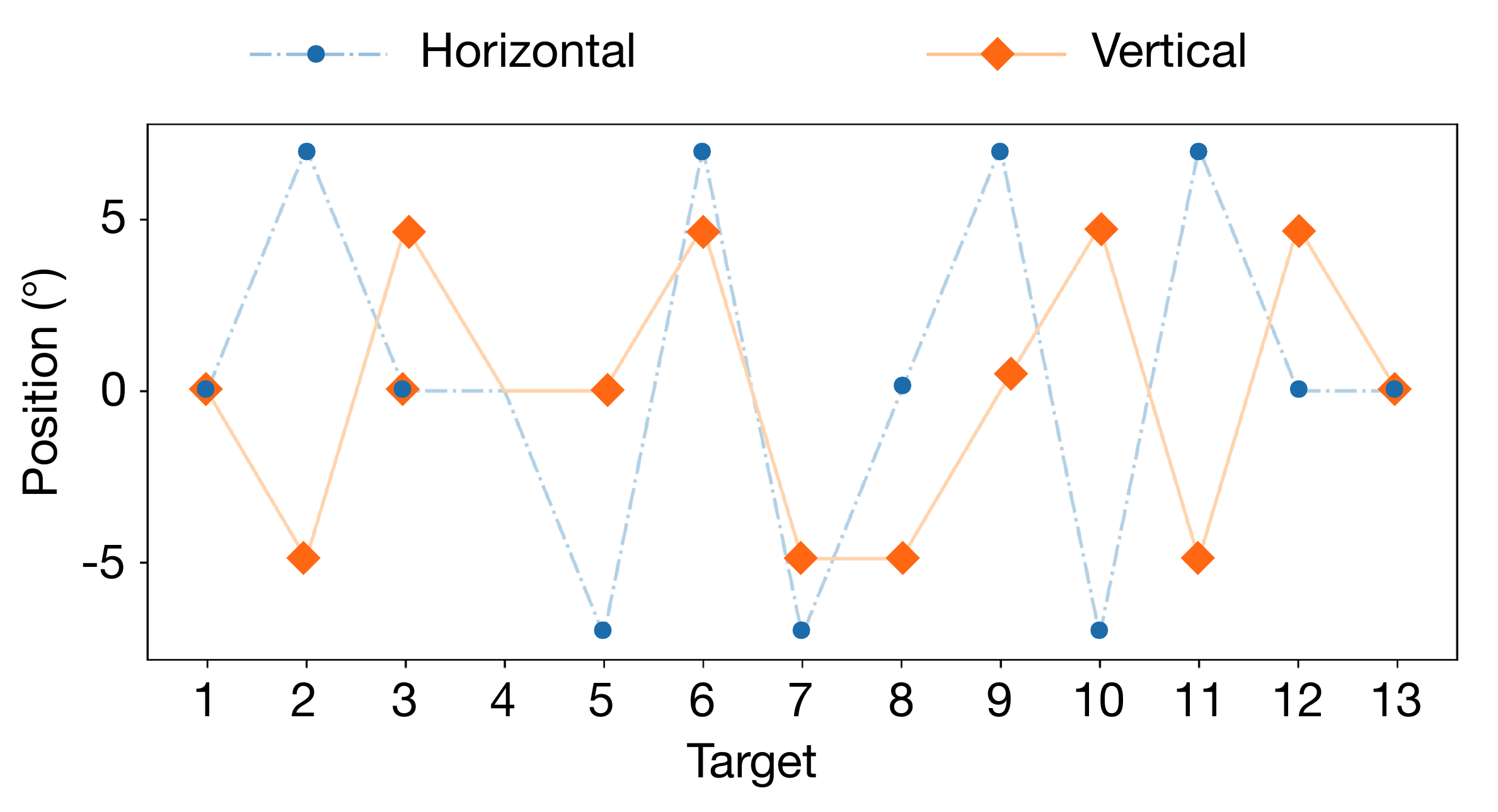}
\end{center}
\caption{Calibration targets designed to maximize variations in gaze angle and directions}
\label{fig:calib}
\end{figure}

\begin{figure*}
\begin{center}
\includegraphics[width=\linewidth]{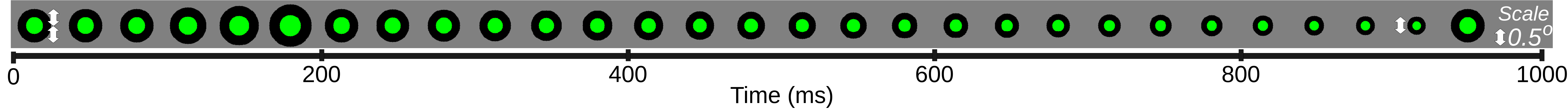}
\end{center}
\caption{Visualization of target size animation over 1-second interval: each frame was displayed for 1/30 second.}
\label{fig:stimuli}
\end{figure*}

Every observer performed a sequence of tasks. Initially, 12 calibration targets were displayed on the screen in a pseudo-random pattern to allow the maximum number of changes in horizontal and vertical directions, as shown in Figure~\ref{fig:calib}. Each target consisted of concentric circles, the larger of which intiially subtended an angle of 1.0 degrees. The target first grew to a size of 1.34 degrees then decreased to a size of 0.5 degrees before disappearing after one second, as shown in Figure ~\ref{fig:stimuli}. The field of view of the calibration targets was 14.19$^{\circ}$ X 9.68$^{\circ}$. The calibration was followed by the tasks described in the following section.

\subsection{Task 1: Calibration Verification Task}
Six calibration verification targets were shown in sequence, subtending a total angle of 10.02$^{\circ}$ X 3.99$^{\circ}$. The delay between the disappearance of one target and the appearance of the next target was on average 31 ms ($\sigma$=5). The calibration verification points were different from those used during calibration.

\subsubsection{Measures}
We evaluated the accuracy and precision with which the methods predicted gaze on the verification targets, assuming that the observer fixated each target. Accuracy measures were based on the difference between the displayed target position and the mean reported gaze position of the stable fixation window. The fixation window was determined from a rolling 450 ms window with a minimum dispersion search after the target was displayed on the screen.

Another essential metric we considered was precision during fixations. Sample-to-sample root mean square error (S2S-RMS) and standard deviation (STD) are two widely used metrics to measure the precision of eye-trackers~\citep[p.~182-4]{edlund2019advanced}. Both measures are related to the spatial variability in the signal over time, but they contain different information about an eyetracker's behavior~\citep[p.~182-4]{edlund2019advanced},~\citep{niehorster2020characterizing}. Because S2S-RMS is calculated on temporally adjacent data points, its value relays information about the spatio-temporal aspects of a system that are absent from STD measures. S2S-RMS is also inherently sensitive to the update rate of the eye-tracker~\citep{blignaut2012precision}. 

\begin{equation}
S2S-RMS =\sqrt{\dfrac{\sum_{i=1}^{n} {(x_{i}-x_{i-1}})^2+(y_{i}-y_{i-1})^2}{n}}
\end{equation}
\begin{equation}
STD=\sqrt{\dfrac{(\sigma_x^2+\sigma_y^2)}{2}} 
\end{equation}

\subsection{Task 2: Smooth Pursuit Task}
To evaluate the eye-tracking methodology, we evaluated two tasks that require very high precision: microsaccade detection and smooth pursuit. For a smooth-pursuit task, observers followed a moving target on a ramp (linear-trajectory) at different velocities (mean=4.6 deg/s, $\sigma$=1.9) with 17 random directional changes.

\subsubsection{Measures}
\label{SP_describe}

Measurement of accuracy and precision in a smooth-pursuit task is not straightforward.~\cite{komogortsev2013automated,komogortsev2010standardization} proposed a method to determine the accuracy based on how closely the smooth-pursuit signal matches the target stimulus. A quantitative smooth pursuit score based on the position and velocity was reported based on the Euclidean distance and differences in speed at every timestamp with respect to the smooth-pursuit target stimulus. 

\begin{figure}[ht]
\begin{center}
\includegraphics[width=\linewidth]{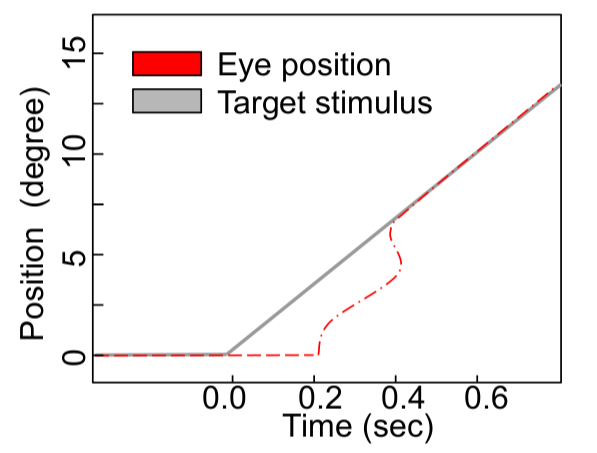}
\end{center}
\caption{Conceptual diagram of eye position (red) and target stimulus (gray) during smooth pursuit after ~\cite{fukushima2013cognitive}.}
\label{fig:sp1}
\end{figure}

We propose a method for measuring precision during smooth pursuit using S2S-RMS and STD after `detrending' the raw data. 'Detrending' a signal subtracts the best-fit line/curve from the data~\cite{moncrieff2004averaging}. So, a constant velocity term can be removed from the smooth-pursuit data resulting in nearly zero-velocity signal. In any smooth pursuit movement with randomly changing directions, it is observed that there is a latency of $100-130$ ms after the direction change~\citep{lisberger1987visual,lisberger1985properties,fukushima2013cognitive}. At that point, the eye either begins moving at approximately the correct velocity (but is lagging the target due to the latency), or the movement starts with a small saccade in the direction of the stimulus. In either case,  the eye velocity is then typically similar to that of the target but lags in position. Finally, any positional offset between the eye and target is corrected by a second `catch-up' saccade, as shown in Figure~\ref{fig:sp1}. At that point, the eye position and velocity match the stimulus. 

In our proposed smooth-pursuit precision metric, we initially find the time interval where both the eye position and eye velocity are closest to the stimulus position and velocity. The eye position at this moment is referred to as the \textit{starting point}. Similarly, we compute the \textit{ending point} just before the stimulus changes direction. An equation describing the line joining the starting and ending gaze points is computed. The gaze signal is detrended using this signal (line), resulting in a signal with zero mean velocity and can be analyzed in the same way as a fixation signal. We can thus compute the precision (S2S-RMS and STD) for the detrended signal.

\subsection{Task 3: Microsaccade Detection Task}
\label{MS_describe}
Motivated by~\citep{shelchkova2018perceptual} and~\citep{Chaudhary_Pelz_2019}, we evoked small, voluntary eye movements with a Snellen acuity chart by displaying a sequence of fixation targets on the teleprompter screen. Any voluntary or involuntary saccade less than 0.5 $\circ$ is considered as \textit{microsaccade} in this paper following ~\cite{engbert2003microsaccades,otero2008saccades,troncoso2008microsaccades,shelchkova2018perceptual,Chaudhary_Pelz_2019}. There were two elements in this task; first, the observer was asked to fixate on a series of thin color bars (6 bars, each (5.2x12 arcmin)), then on six colored boxes, alternating in size between 12x12 and 5.2x12 arcmin. Each target was displaced from the previous target in the horizontal direction by 0.2 degrees (12 arcmin). The total expected number of small eye movements was ten (five each for the color bars and boxes).

\subsubsection{Measures}
We measured the number of microsaccades detected when the observer looked to the different color bars$/$boxes. Horizontal eye movements detected within 100 - 500 ms of each target onset were identified as microsaccades. Note that horizontal cyclopean velocity was used for microsaccade identification. We used the method described in~\citep{Chaudhary_Pelz_2019}, where the velocity signal is filtered with a 1D total variational denoising filter (regularization value of 0.05). After denoising the velocity signal, a threshold value is determined with an adaptive algorithm by fitting two velocity distributions representing noise and microsaccade based on Gaussian mixture models. Velocities above the adaptive threshold are identified as microsaccades based on the Velocity-Threshold Identification algorithm (I-VT)~\citep{salvucci2000identifying}. We refer the reader to~\citep{Chaudhary_Pelz_2019} (Section: Microsaccade detection) for additional details. 

\section{Results}
\subsection{Iris segmentation}
Maintaining the aspect ratio by cropping to 540$\times$540 before resizing to 224$\times$224 boosted the performance by 1.8\% and 2.9\% for the uncorrelated and correlated test data respectively as seen in Table~\ref{tab:1}.
\begin{table}[ht]
\caption{Note improvement in segmentation results due to change in aspect ratio}
\label{tab:1}       
\begin{tabular}{|c|c|c|}
\hline
\textbf{Dataset} &~\cite{Chaudhary_Pelz_2019} & Ours\\
\hline
Training & 89.8\% & 92.4\% \\
\hline
Uncorrelated test & 89.1\% & 90.7\%  \\
\hline
Correlated test  & 86.6\% & 89.1\%\\
\hline
\end{tabular}
\end{table}

\subsection{Qualitative Results}
\label{qualitative}
\begin{figure}[h]
\begin{center}
\includegraphics[width=\linewidth]{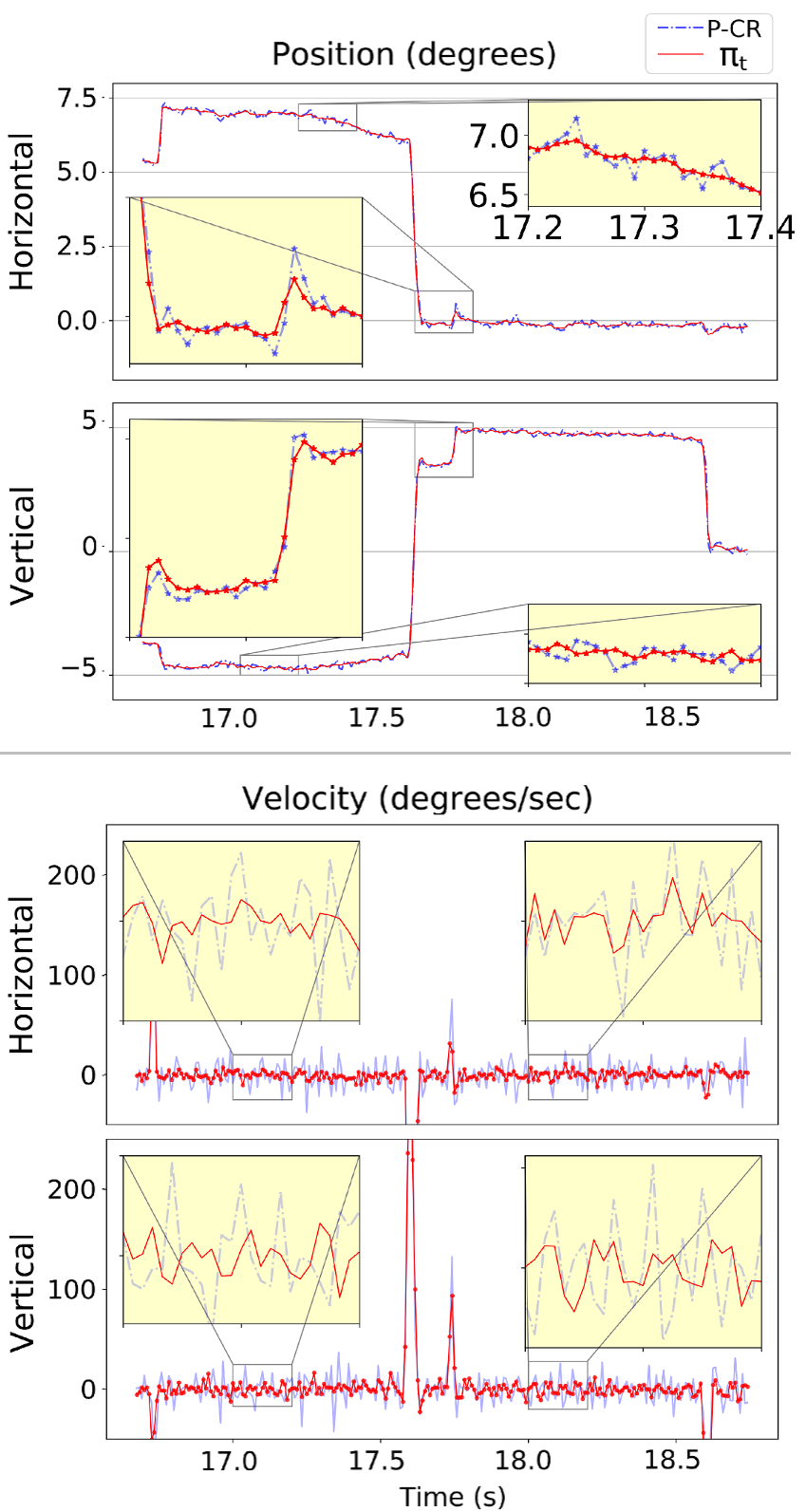}
\end{center}
   \caption{Comparison of position (top panel) and velocity (bottom panel) signals from P-CR and $\pi_t$ methods in horizontal and vertical directions (best viewed on online version of the paper). Each zoomed-in section (in yellow) is 200 ms in duration (4$\times$).} In zoomed-in section (*) marker shows estimate of individual timestamp and represents similar latency to P-CR based method.
\label{fig:long_PCR}
\end{figure}

The dramatic difference in noise between the traditional P-CR and $\pi_t$ methods can be seen in Figure~\ref{fig:long_PCR}. The top panel shows the horizontal and vertical position signals for both methods; the inset panels show the indicated segments at $4\times$ magnification. Note that the noise reduction from the $\pi_t$ method does not introduce a temporal lag in the signal as temporal filtering methods do (distinguishable with (*) marker in position plots). The lower panel shows the velocity signals over the same periods for both methods.  The reduced noise inherent in the $\pi_t$ method is especially evident in the velocity signals. Note that all the results presented for traditional P-CR and $\pi_t$ methods are analyzed for same video at the same timestamps.

\subsection{Task 1: Verification Task Performance}
\subsubsection{Accuracy}
\begin{figure*}[h]
\begin{center}
\includegraphics[width=\linewidth]{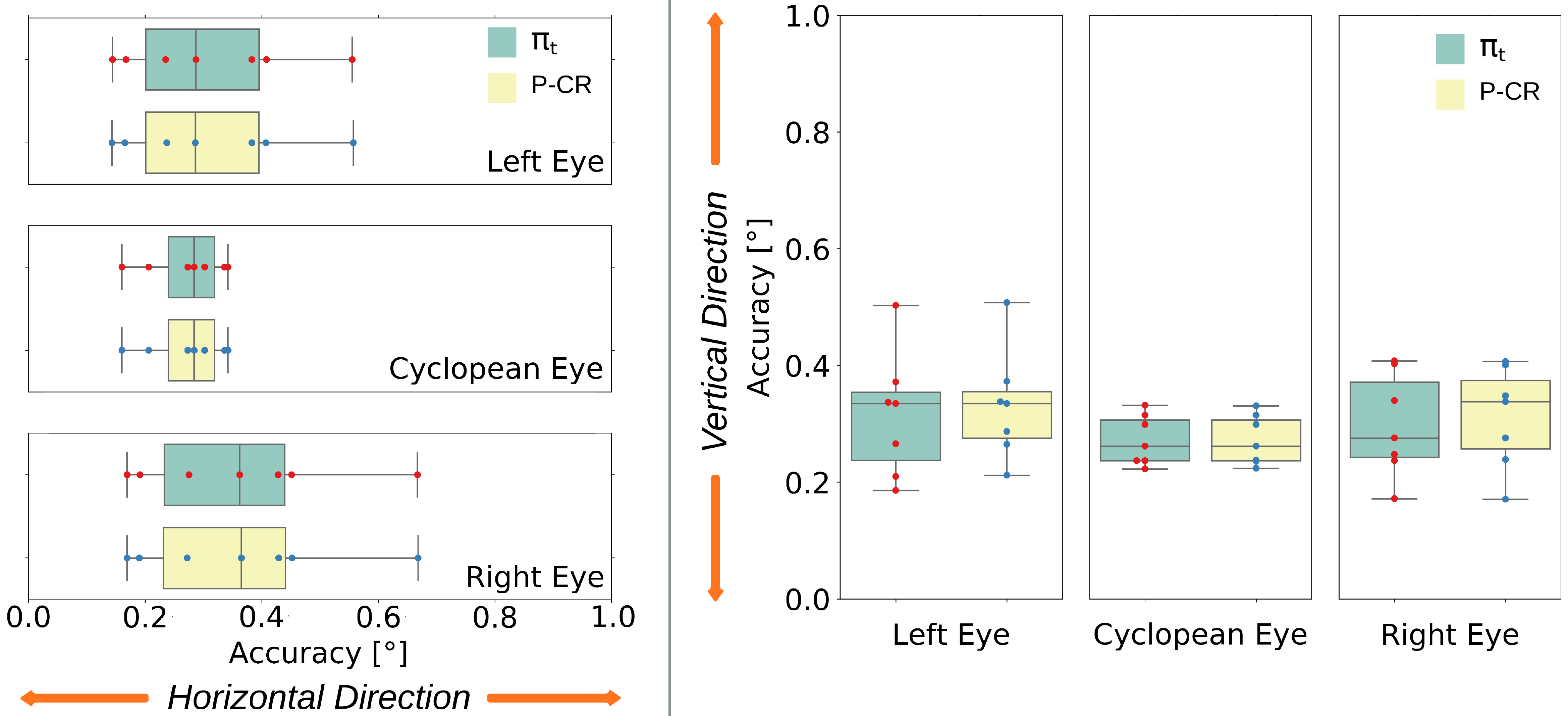}
\end{center}
\caption{Box plot indicating \textbf{accuracy} in the horizontal and vertical direction using $\pi_t$ and P-CR based methods for left, right, and cyclopean eyes. Similar accuracy values indicate drifts are addressed in $\pi_t$ based method.}
\label{fig:Accxverify}
\end{figure*}

Figure~\ref{fig:Accxverify} (left) shows the accuracy of $\pi_t$ and P-CR methods along the horizontal axis. Both plots indicate accuracy for the left, right, and cyclopean eye. Each point in the figure represents an individual participant. Figure~\ref{fig:Accxverify} (right) shows the same data for accuracy along the vertical axis. Note that accuracy is the same for $\pi_t$ and P-CR based methods.

\subsubsection{Precision}
\begin{figure*}[h]
\begin{center}
\includegraphics[width=\linewidth]{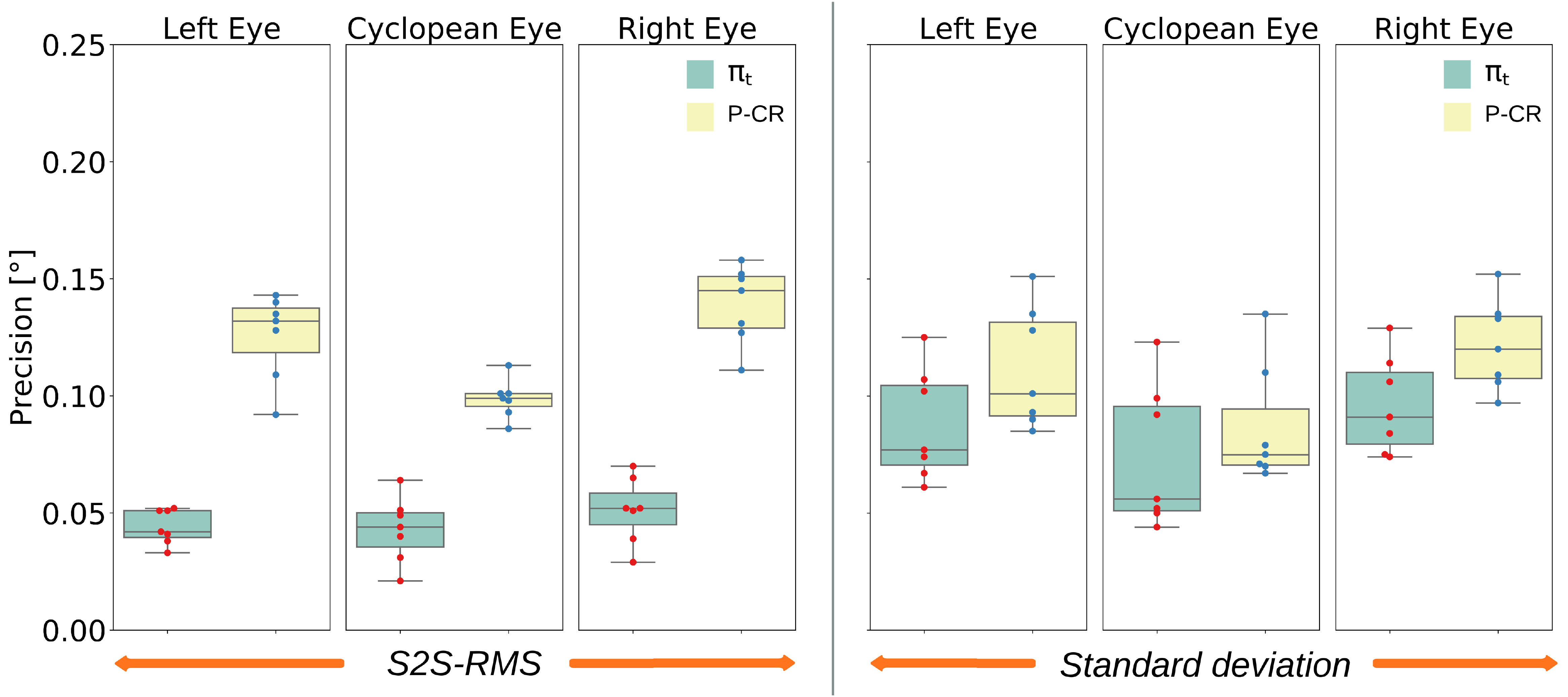}
\end{center}
\caption{Box plot indicating \textbf{precision} (S2S-RMS and standard deviation) using $\pi_t$ and P-CR based methods for left, right, and cyclopean eyes.}
\label{fig:STDPrecisio}
\end{figure*}

While the accuracy values are the same, the $\pi_t$ method has significantly better precision. Figure~\ref{fig:STDPrecisio} represents the precision for $\pi_t$ and P-CR based methods measured with S2S-RMS (left panel) and standard deviation (STD) (right panel) metrics. Each figure has results for individual left and right eyes and for the combined, cyclopean eye. 
The median sample-to-sample root mean square (S2S-RMS) and the standard deviation value are improved by at least 55\% and 23\%, respectively.

\subsection{Task 2: Smooth Pursuit Performance}
\begin{figure}
\begin{center}
\includegraphics[width=\linewidth]{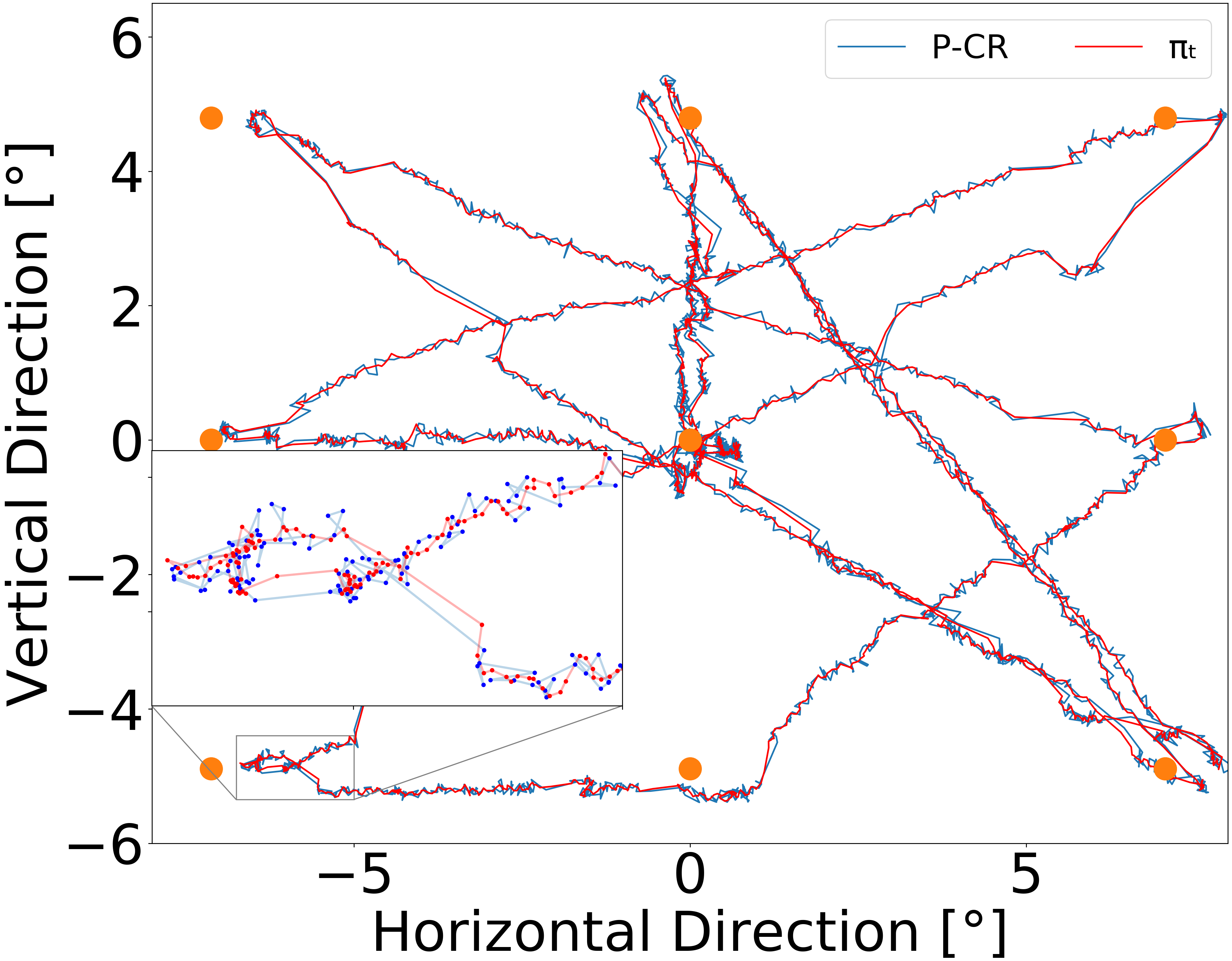}
\end{center}
\caption{Gaze map of eye movements during a smooth pursuit task for one participant. The red and blue traces show movements calculated with $\pi_t$ and P-CR methods, respectively. The orange points indicate the extreme position of the target for each direction change. Rectangular box indicates the zoomed in section (4$\times$) (best viewed on online version of the paper).}
\label{fig:sp_display}
\end{figure}

\begin{figure*}[h]
\begin{center}
\includegraphics[width=\linewidth]{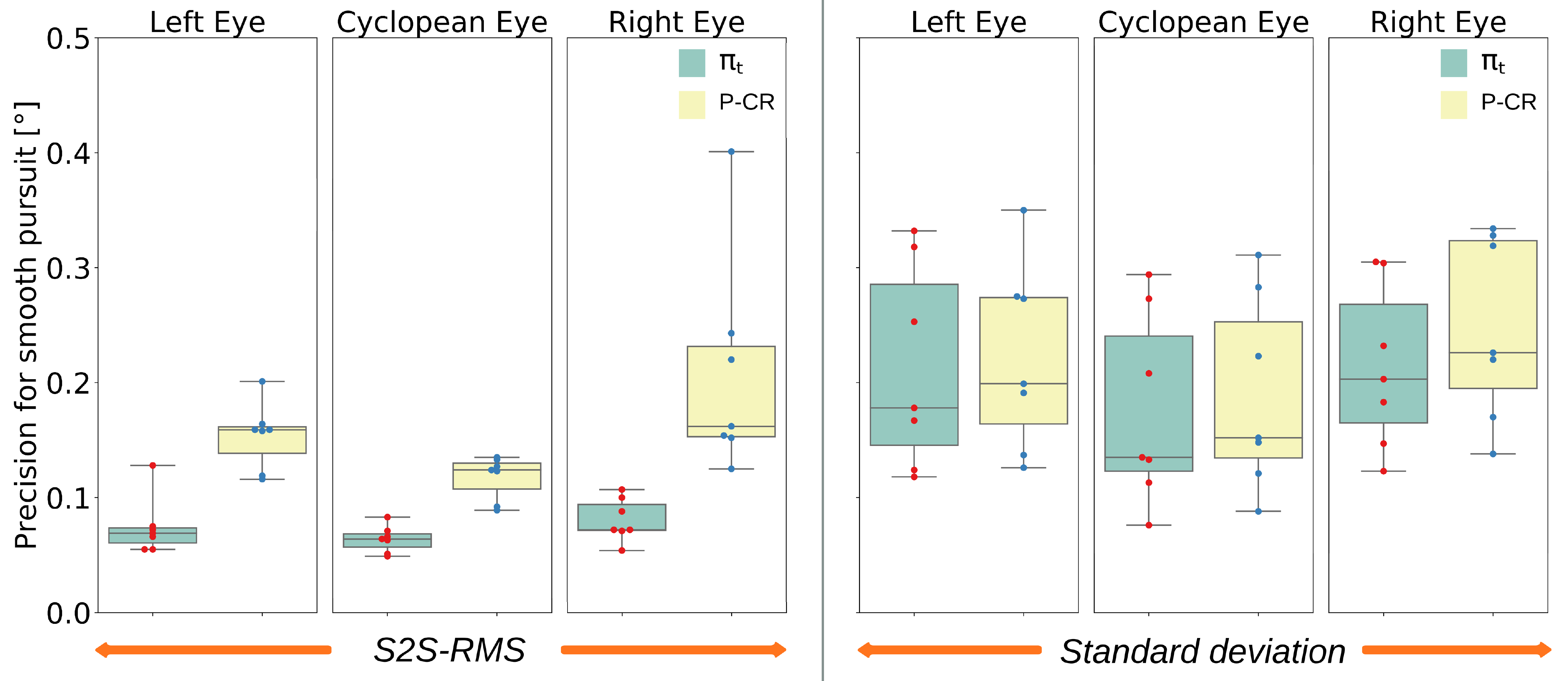}
\end{center}
\caption{Box plot indicating  \textbf{precision} (S2S-RMS and standard deviation)  \textbf{using detrended smooth pursuit} signal obtained from $\pi_t$ and P-CR based methods for left, right, and cyclopean eyes.}
\label{fig:sp_std}
\end{figure*}

Figure~\ref{fig:sp_display} represents the gaze map of eye movements during a smooth pursuit task for one participant. Note that small, high-frequency fluctuations during smooth pursuit are minimized in the $\pi_t$ record compared to P-CR. Figure~\ref{fig:sp_std} represents the precision for $\pi_t$ and P-CR based methods using S2S-RMS (left) and STD (right) metrics. Each figure has results for individual left and right eye and the combined cyclopean eye. Note that the reported precision is for the detrended signal, as described in Section~\ref{SP_describe}. The median sample-to-sample root mean square (S2S-RMS) and the standard deviation value are improved by at least 48\% and 10\%, respectively.

\subsection{Task 3: Microsaccade Detection Performance}
\begin{figure*}
\begin{center}
\includegraphics[width=\linewidth]{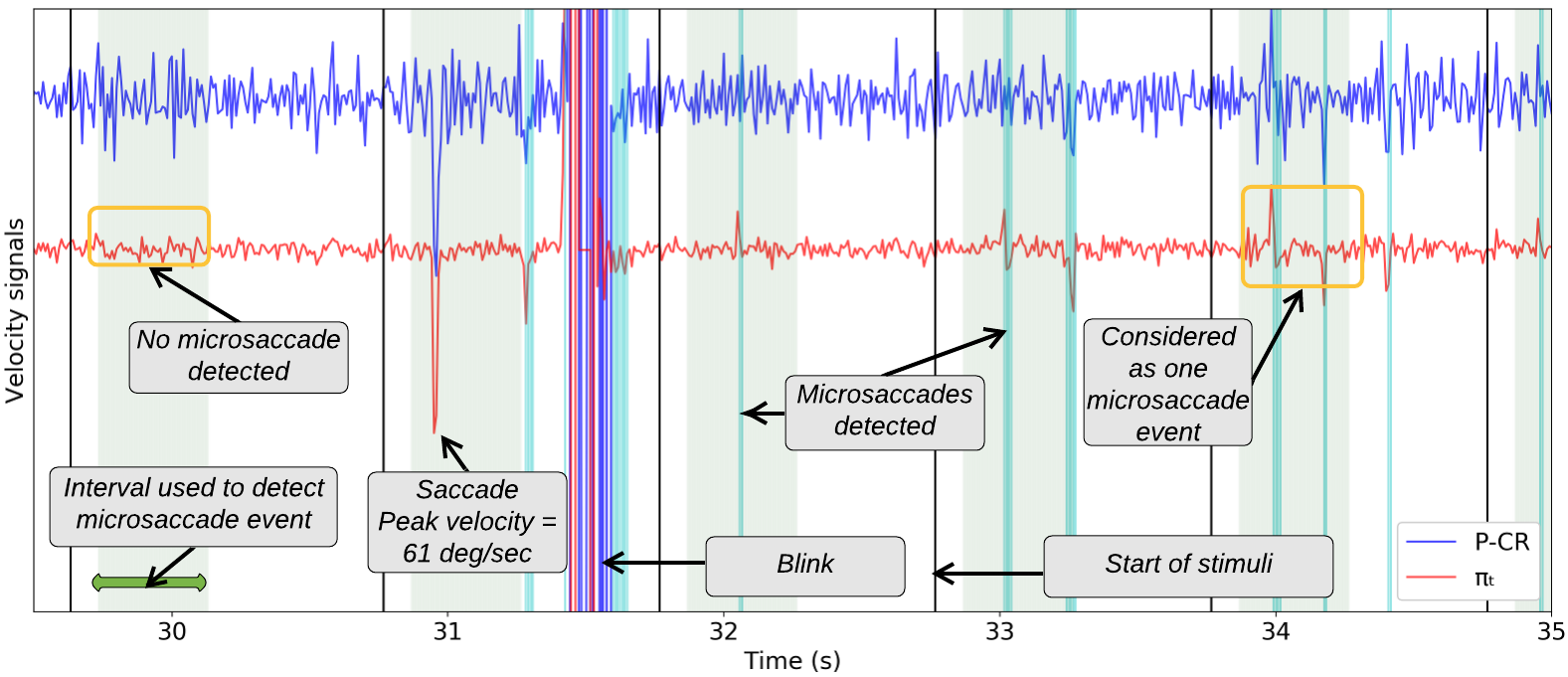}
\end{center}
\caption{Figure shows the velocity signals for P-CR and the $\pi_t$ model. Each microsaccade event starts with a display of a stimuli (indicated in vertical black line). The time period where an event is likely to occur is indicated in green and the detected microsaccade is indicated in cyan. The yellow boxes indicate special events when microsaccades are not detected, or multiple microsaccade detected in the time interval (best viewed on online version of the paper).}
\label{fig:microsaccade}
\end{figure*}

As described in section~\ref{MS_describe}, out of ten possible small microsaccades per subject, the number of microsaccades detected by the seven subjects was [10, 8, 8, 7, 7, 6, 5]. Only the first detected microsaccade event during a time interval of 100-500 ms of each target onset was counted. If no small-movements were beyond the threshold value during that time interval, then the count was zero. Note that 27\% of the microsaccades are not detected, mainly because of the significant head movement and simultaneous eye movement of the person to fixate at the gaze position. Figure~\ref{fig:microsaccade} shows a number of microsaccades detected with the $\pi_t$ model.  The P-CR signal has too much noise to allow the microsaccades to be detected.

\subsection{Simulation Test}
In this section, we explore the mathematical formulation in Equation~\ref{eq:main2} using simulated data. Our hypothesis is, given any two random signals ($A_{noise}$ and  $B_{drift}$) which are related to each other by first-order derivative/integration; where signals $A_{noise}$ and  $B_{drift}$ are influenced by spatial noise and temporal drift respectively, then the $\pi_t$ algorithm can be used to estimate a signal that compensates for both spatial noise and temporal drift. Note that rather than attempting to replicate eye-gaze data we are simulating random signals $A_{noise}$ and  $B_{drift}$ that fulfill the stated requirements. 

\begin{figure*}
\begin{center}
 \includegraphics[width=\linewidth]{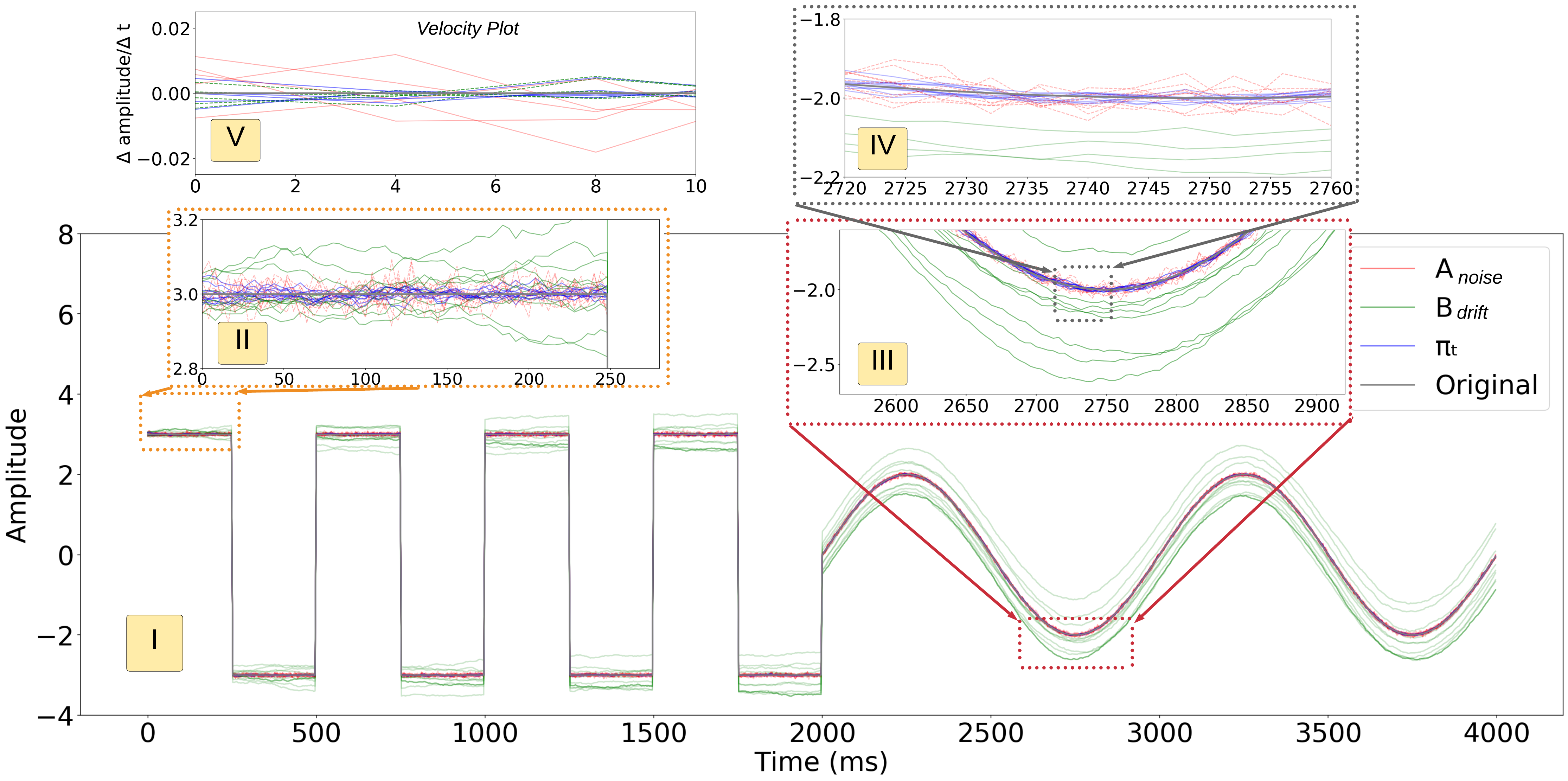}
\end{center}
\caption{The figure shows the original signal (solid grey line), ten trials of signals $A_{noise}$ and $B_{drift}$, and its recovered $\pi_t$ output. Note that signal $A_{noise}$ has spatial noise (dashed red lines visible in block II indicated in a yellow box), signal $B_{drift}$ has temporal drift (solid green lines visible in block III indicated in a yellow box). The derived signal output (blue lines, distinguishable in zoomed-in blocks IV (position signal) and V (velocity signals) handles the temporal drift and minimizes the spatial noise (best viewed on online version of the paper).}
\label{fig:simulation}
\end{figure*}

Signals $A_{noise}$ and  $B_{drift}$ are derived from a 2 Hz square wave with an amplitude of three units sampled at 250 Hz for 2 sec followed by a 1 Hz sine wave with a peak amplitude of two units sampled at 250 Hz for the next two seconds. Signal $A_{noise}$ consists of the addition of random Gaussian noise ($\mathcal{N}(0,0.03)$ with a sampling size of 1000) to the \textit{original signal} (position-like signal; noisy). Signal $B_{drift}$  consists of the addition of random Gaussian noise ($\mathcal{N}(0,0.01)$) in the \textit{spatial gradient of the original signal} (velocity-like signal; temporal drift). Figure \ref{fig:simulation} shows ten such trials and its derived output based on $\pi_t$.

For quantitative results, Table \ref{tab:sim} shows the performance of our approach $\pi_t$ generated using these two signals ($A_{noise}$ \& $B_{drift}$) in terms of mean square error (MSE) and $R^2$ computed against the original signal for 100 trials. Note MSEo $R^2$o represents computation of these metrics in the original signal domain,  MSEt $R^2$t represents computation with the gradient of the signal. We observe an improvement in the original domain signal as well as a gradient-domain signal for both metrics MSE, and $R^2$ for our $\pi_t$ derived signal. 
\begin{table}[h]
\caption{Simulation test; MSEo $R^2$o represents computation of these metrics in the original signal domain,  MSEt $R^2$t represents computation with the gradient of signal. MSEt and MSEo are in $*10^{-4}$ units.  }
\label{tab:sim}       
\begin{tabular}{|c|c|c|c|}
\hline
  & Signal $A_{noise}$ & Signal $B_{drift}$ & $\pi_t$  \\
\hline
MSEo & 9.04 $\pm$ 0.38 & 475.71 $\pm$ 480.30 & \textbf{1.51 $\pm$ 0.12}\\
\hline
MSEt & 18.12 $\pm$ 1.06 & 1.0 $\pm$ 0.04 & \textbf{0.88 $\pm$ 0.04}\\
\hline
$R^2$o & 0.9998  & 0.9914  & \textbf{0.9999}  \\
\hline
$R^2$t & 0.9931 & 0.9996  & \textbf{0.9997}  \\
\hline
\end{tabular}
\end{table}

\section{Discussion}
The previous methods based on tracking motion of iris features offer significant improvements over existing methods, but an important limitation of these methods is the temporal drift in the estimation of gaze.  The major contributing factor for the drift was the compounding of small errors in the approximation of velocity.  Our work addresses this issue by considering the pupil edges (extensively used in traditional gaze tracking methodologies) as support for velocity approximation. We present a new mathematical formulation which handles drift in gaze position by considering a weighted sum in a Kalman filter framework.

Moreover, our approach is being extended further such that we will also be capable of handling cases during blinks and insufficient texture matches. During calibration, the use of positional information from $\pi_t$ helps to eliminate the need for integration of velocity, which was the bottleneck during blinks in~\citep{Chaudhary_Pelz_2019, pelzwitzner}. Our modification in $\beta_P$ and $\beta_I$, as described in section ~\ref{sec:blinks}, helps to handle cases of blinks and insufficient texture matches. Note that in cases of partial blinks, complete pupil edges are often not visible, and ellipse fits cannot be computed reliably. In such scenarios, there is still a possibility of a good number of feature matches. The proposed method allows for the study of eye movements even during the period immediately before and after a full blink, which leads to errors when relying on ellipse fits. Further, with traditional methods, there is a high possibility of ellipse fit on false edges, which is eliminated in our method as we rely on a large number of iris features.

Our mathematical formulation can further be interpreted as the superposition of contributions from separate pupil and iris estimates. In this case, we consider the signal as the combination of a high-temporal frequency component from the iris motion vectors and low-temporal frequency components from pupil position. This combination helps to maintain the accurate low-frequency position signal while supporting high precision tasks based on the high-frequency component from the stable iris motion vectors. This formulation eliminates reliance on high-frequency components from the pupil position, which are noisy and interfere in the study of eye movements such as smooth pursuit and microsaccades.

 Our method of calibration of the two components separately is critical. This is because the iris position drifts over time, and we need the same scaling factor for the combination. A mismatch in the scaling factor creates undershoot or overshoot during saccades as the mathematical formulation is the weighted sum of two components when the low-frequency component is significant. Thus, we propose a new iris calibration technique based solely on the relative change in iris velocity with the change in calibration targets.

The hybrid  ($\pi_t$) approach shows a comparable accuracy with the P-CR based approaches, suggesting that drift is not prevalent in our method even when iris information is incorporated. The simulation results in Table ~\ref{tab:sim} show a low value of MSEo and MSEt in $\pi_t$ compared to original individual signals $A_{noise}$ and  $B_{drift}$, suggesting that this work can be applied in any domain where noise is prevalent in one signal and temporal drift in another.  The small values in MSEo verify our contribution in handling temporal drift existing in $\pi_t$  based methods.

The advantage of our ($\pi_t$) method is the improvement in the precision values without affecting the accuracy in the system. An important performance metric for eye-tracking systems is S2S-RMS, as it is highly influenced by the frame rate and temporal variation. Our method also shows an improvement in median value of S2S-RMS with at least a 55\% reduction (0.132$^{\circ} \rightarrow$ 0.042 $^{\circ}$, 0.145$^{\circ} \rightarrow$ 0.052$^{\circ}$, and 0.099 $^{\circ} \rightarrow$ 0.044 $^{\circ}$ for left, right and cyclopean eyes) in the verification task for the same video sequences. Further, we have shown an improvement of 23\% in STD, which can be thought of as a combination of both eye movement and eye-tracking methodology noise. 

Additionally, we show an improvement of at least 48\% in S2S-RMS and 10\% in STD for smooth pursuit tasks, demonstrating the value of the method for studies of smooth pursuit. We also highlight an essential contribution of~\citep{Chaudhary_Pelz_2019}'s previous work in detecting eye movements as small as 0.2$^{\circ}$ and also verifies that this formulation does not deteriorate the eye gaze signal quality, as all the missing detection were because of simultaneous head and eye movements to fixate at the target. Lowering the regularization value (section ~\ref{MS_describe}) helps in detection of few of these movements but it increases the number of false alarms. A relevant field of application to this strategy would be the study of vergence eye movement as we have boosted the ability of eye trackers to study movements with small changes.

Our study improves the precision of current video-based eye trackers by relying on multiple features of the human eye. The method is not without limitations, of course. First, it requires multiple calibration points in order, which include a series of changes in X-Y direction (refer to Figure~\ref{fig:calib}). It will require an extra effort for calibration design, though the present study does not answer the minimum calibration point changes required for proper iris calibration; we used 12 changes, and addressing the minimum number is recognized as future work. Second, the performance of our method falls in the case of blurred images as we rely on iris features for the velocity signal and on pupil edges for pupil position estimates. Lack of a sharp eye image is also a problem for many current video-based systems as well, but this can be solved with a higher-resolution camera and proper exposure time. Third, finding the precision matrix in Equation ~\ref{eq:3conv} is a time-consuming step that increases as the time series increase. For our short videos, it was not an impediment, but for general use using long series, we can incorporate a precision matrix using techniques like LU-decomposition~\citep{banachiewicz1938methode} or use small windowing blocks to speed processing. Finally, compression artifacts are still a problem in our method as mentioned in~\citep{Chaudhary_Pelz_2019}

In summary, the main contributions of this paper are handling of spatial drift, a novel strategy that incorporates the two signals in different domains that are related by first-order derivative/integration (one noisy the other with temporal drift), a new calibration routine for iris calibration, a way to compute precision in smooth pursuit signals by signal detrending, and most importantly, a method for estimating gaze that allows for significant improvement of precision compared to current video-based eye-tracking methodology. At the same time, this method is also useful in tasks requiring high precision, such as the study of microsaccades, smooth pursuit, and even vergence eye movements.

This methodology can be used with any eye tracker (P-CR, 3D based approaches, appearance-based models) that estimate the head compensated gaze position to improve its precision. It also gives a confidence value for each signal component, supporting further study and error analysis.

\section{Acknowledgements}
The authors want to acknowledge the contribution of our high school intern, Brian Cowburn, for helping hardware setup for synchronizing iPad display and the IRLED with a photo-diode setup.

\newpage
\bibliographystyle{ACM-Reference-Format}
\bibliography{ref}  

\end{document}